\documentclass[sigconf, table]{acmart}

\usepackage{array}
\usepackage{graphicx}
\usepackage{listings}
\usepackage{booktabs} 
\usepackage{multirow} 
\usepackage{subcaption}
\usepackage{verbatim} 
\usepackage{mathtools}
\usepackage[linesnumbered,ruled,vlined]{algorithm2e}
\usepackage{color}

\usepackage{flushend}

\usepackage{soul}

\mathtoolsset{firstline-afterskip=0pt}

\colorlet{dark-green}{green!70!black}

\newcommand{\down}{{\color[rgb]{0.7,0,0}{$\blacktriangledown$}}}
\newcommand{\up}{{\color[rgb]{0,0.45,0.1}{$\blacktriangle$}}}
\newcommand{\eq}{{\color[rgb]{0,0,0.5}{$\blacklozenge$}}}

\newcolumntype{A}{>{\raggedright\arraybackslash}m{1.65cm}}
\newcolumntype{B}{>{\raggedright\arraybackslash}m{1.75cm}}

\begin{document}
	
	\title[Solving the Exponential Growth of Symbolic Regression Trees in GSGP]{Solving the Exponential Growth of Symbolic Regression Trees in Geometric Semantic Genetic Programming} 
    
	\author{\Large Joao Francisco B. S. Martins, Luiz Otavio V. B. Oliveira, Luis F. Miranda, Felipe Casadei, Gisele L. Pappa}
	\affiliation{%
		\institution{Universidade Federal de Minas Gerais, Department of Computer Science}
		\city{Belo Horizonte} 
		\country{Brazil}
	}
	\email{[joaofbsm, luizvbo, luisfmiranda, casadei, glpappa]@dcc.ufmg.br}
	\renewcommand{\shortauthors}{Martins et al.}
	
	\begin{abstract}
		
		Advances in Geometric Semantic Genetic Programming (GSGP) have shown that this variant of Genetic Programming (GP) reaches better results than its predecessor for supervised machine learning problems, particularly in the task of symbolic regression. 
		However, by construction, the geometric semantic crossover operator generates individuals that grow exponentially with the number of generations, resulting in solutions with limited use.
		This paper presents a new method for individual simplification named GSGP with Reduced trees (GSGP-Red). 
GSGP-Red works by expanding the functions generated by the geometric semantic operators. The resulting expanded function is guaranteed to be a linear combination that, in a second step, has its repeated structures and respective coefficients aggregated.		
		Experiments in 12 real-world datasets show that it is not only possible to create smaller and completely equivalent individuals in competitive computational time, but also to reduce the number of nodes composing them by 58 orders of magnitude, on average.
		
	\end{abstract}
	
	%
	%

	\ccsdesc[500]{Computing methodologies~Genetic programming}
	\ccsdesc[300]{Computing methodologies~Supervised learning by regression}
	
	
	\keywords{Genetic Programming; Geometric Semantic Genetic Programming; Symbolic Regression; Solution Size; Function Simplification}
    
    \copyrightyear{2018} 
    \acmYear{2018} 
    \setcopyright{acmlicensed}
    \acmConference[ GECCO '18]{Genetic and Evolutionary Computation Conference}{July 15--19, 2018}{Kyoto, Japan}
    \acmPrice{15.00}
    \acmDOI{10.1145/3205455.3205593}
    \acmISBN{978-1-4503-5618-3/18/07}

	\maketitle
	
	\section{Introduction}

The Geometric Semantic Genetic Programming (GSGP) \cite{moraglio2012geometric} framework introduces geometric semantic operators to Genetic Programming (GP). These operators are capable of inducing a semantic effect through syntactic operations, and allow GSGP to explore a conic semantic fitness landscape, which can be efficiently optimized using evolutionary search \cite{moraglio2011abstract}. GSGP has shown to outperform GP in different scenarios, specially in symbolic regression \cite{castelli2013prediction, castelli2014prediction, vanneschi2014genetic}.

However, GSGP suffers from a problem that limits its use in real-world applications. By definition, the geometric semantic operators generate offspring composed by the complete representation of their parents, plus some additional structures, leading to an exponential growth of the solution size in the number of generations  \cite{vanneschi2017introduction}. This extreme growth leads to excessive usage of memory and computational power, and also results in non-interpretable solutions \cite{vanneschi2017introduction}. 

Two different approaches have been followed in the literature to deal with the problem of exponential growth of GSGP individuals, although they were not able to effectively solve the problem. The first approach simply focuses on making the algorithm more efficient in terms of memory and computational resources \citep{moraglio2014efficient,vanneschi2013new}. The second proposes new versions of the semantic crossover operators since they are the ones responsible for the excessive growth of the solutions \citep{pawlak2017competent,nguyen2016subtree}.

Most works based on GSGP follow the first approach, using an implementation
presented by \citet{vanneschi2013new} that stores pointers to the trees representing the individuals, instead of keeping the whole individual in memory. This implementation computes the semantics---and fitness---of new individuals from the values of their parents \cite{castelli2014cpp}. Although the implementation is very fast and reduces the memory needed during the evolution, the individuals are not explicitly built during the search. Thus, if we want to access the final individual or if new data is presented after the training stage, an assembling step is needed to generate the complete individual from the pointers, which still presents an exponential size. 

This work presents a new method, called Geometric Semantic Genetic Programming with Reduced trees (GSGP-Red), to solve the problem of excessive growth of GSGP solutions for symbolic regression. GSGP-Red works by expanding the functions generated by the geometric semantic operators. The resulting expanded function is guaranteed to be a linear combination that, in a second step, has its repeated structures and respective coefficients aggregated. These expansion and aggregation operations ensure that only one copy of each function composing the solution is kept in the simplified individual, leading to a massive reduction of the size of the resulting solutions while guaranteeing the exact same results of GSGP.

An experimental analysis comparing the proposed method with GSGP and GP in 12 real-world datasets showed that GSGP-Red is capable of finding solutions equivalent to the ones generated by GSGP while resulting in individuals up to 64 orders of magnitude smaller than those generated by previous GSGP versions, with a practicable additional overhead in computational time.

The remainder of this paper is organized as follows. Section 2 introduces the main concepts of GSGP. Section 3 reviews related work, while Section 4 introduces the proposed method. Section 5 reports computational results, and Section 6 draws conclusions and points out direction of future work.
	\section{Geometric Semantic Genetic Programming}

Genetic Programming (GP) \cite{koza1992genetic} manipulates individuals during the evolution by applying operators that modify the structure of their trees, i.e., their syntax. Although some restrictions are respected by GP operators---e.g., the arity of the function nodes---they do not consider the behaviour---i.e., the semantics---of the individuals. GSGP \cite{moraglio2011abstract}, on the other hand, employs operators that act on the syntax of the population with a defined semantic outcome.

In the context of symbolic regression, the semantics of a given individual $p$, representing a symbolic expression---usually stored as a tree---can be represented as the output vector it generates when applied to the training set $T=\{(\mathbf{x}_i,y_i)\}_{i=1}^n$---with $(\mathbf{x}_i,y_i) \in \mathbb{R}^d\times \mathbb{R}$ for $i=1,2,\ldots,n$---given by $s(p)=[p(\mathbf{x}_1), p(\mathbf{x}_2), \ldots, p(\mathbf{x}_n)]$. This representation allows us to describe the semantics of any individual in an $n$-dimensional semantic space \cite{oliveira2016dispersion}. 

GSGP defines geometric semantic operators that generate offspring with a given behaviour in the semantic space w.r.t. a given metric. Given a parent individual $p$, the Geometric Semantic Mutation (\textit{GSM}) operator applies a semantic perturbation to the individual, generating an offspring placed inside a ball centred on the parent individual in the semantic space, with radius $\varepsilon\in \mathbb{R}$ proportional to the mutation step. The operator is defined as

\begin{equation}\label{eq:gsm}
	\textit{GSM}(p(\mathbf{x}), \delta) = p(\mathbf{x})+\delta\times(r_m(\mathbf{x})-r_n(\mathbf{x}))~,
\end{equation}

\noindent
where the parameter $\delta$ is the mutation step and $r_m$ and $r_n$ are functions randomly built.

The Geometric Semantic Crossover (\textit{GSX}) operator, on the other hand, combines two individuals $p_1$ and $p_2$, resulting in a single offspring placed in the metric segment connecting the parents in the semantic space. The \textit{GSX} operator defined w.r.t. the Euclidean distance is given by 

\begin{equation}\label{eq:gsx-e}
	\textit{GSX}_E(p_1(\mathbf{x}), p_2(\mathbf{x})) = k\times p_1(\mathbf{x}) + (1-k)\times p_2(\mathbf{x})~,
\end{equation}

\noindent
where $k\in \mathbb{R}$ is a constant uniformly sampled from $[0,1]$. Similarly, the GSX operator defined w.r.t. the Manhattan distance is given by

\begin{equation}\label{eq:gsx-m}
	\textit{GSX}_M(p_1(\mathbf{x}), p_2(\mathbf{x})) = r_f(\mathbf{x}) \times p_1(\mathbf{x}) + (1-r_f(\mathbf{x}))\times p_2(\mathbf{x})~,
\end{equation}

\noindent
where $r_f$ is a function randomly generated with codomain $[0,1]$.

By construction, the geometric semantic mutation and crossover operators induce, respectively, linear and exponential growth of the individuals with the number of generations. Equations \ref{eq:avg-size-GSM}, \ref{eq:avg-size-GSX-E} and \ref{eq:avg-size-GSX-M} present the expected number of nodes of an individual of the generation $g > 0$, generated by GSGP using only one of the geometric semantic operators---\textit{GSM}, \textit{GSX}$_E$ or \textit{GSX}$_M$, respectively \cite{pawlak2015thesis, oliveira2016improving}. $E[P_0]$ is the expected number of nodes in the individuals of the initial population, $E[r]$ is the expected number of nodes in the random functions generated by the operators and $a, b$ and $c$ are the number of additional nodes (constant) used by $\textit{GSM}$, $\textit{GSX}_E$ and $\textit{GSX}_M$, respectively.

\begin{equation}
	\label{eq:avg-size-GSM}
	E[\textit{GSM}, g]=E[P_0]+g\times (2\times E[r]+a)
\end{equation}

\begin{equation}
	\label{eq:avg-size-GSX-E}
	E[\textit{GSX}_E, g]=2^{g}\times E[P_0]+(2^{g}-1)\times b
\end{equation}

\begin{equation}
	\label{eq:avg-size-GSX-M}
	E[\textit{GSX}_M, g]=2^{g}\times E[P_0]+(2^{g}-1)\times (E[r]+c)
\end{equation}

This characteristic is pointed out as the main drawback of GSGP---after a few generations the population becomes unmanageable in terms of memory and computational time spent to compute the fitness \cite{vanneschi2017introduction}. In addition, the excessive size of the individuals makes the functions they represent very hard to understand and interpret \cite{castelli2016controlling}. Since one of the main advantages of GP over other black box learning approaches is the ability to find solutions in the form of comprehensible structures, the exponential size of GSGP solutions limits its usage in practice \cite{nguyen2016subtree}.

	\section{Related Work}

The exponential growth of GSGP individuals was identified by \citet{moraglio2012geometric} in their seminal work. The authors propose to simplify the offspring during the evolution in order to keep the size of the individuals manageable. However, given the complexity of the process, they suggest simplifying the functions only sufficiently---i.e., partially instead of optimally---using, for example, a computer algebra system in order to avoid increasing the computational cost of GSGP excessively. 

There are a few works in the literature that try to deal with the problem of tree exponential growth, but none of them actually solve it. These methods follow two main directions. The first focuses on more efficient implementations of GSGP, while the second proposes different modifications to the genetic operators aiming to reduce the size of the produced offspring---resulting in operators that are only approximately geometric.

Among works in the first group are the implementation of \textit{geometric semantic operators for symbolic regression}
proposed by Vanneschi and colleagues \cite{vanneschi2013new,castelli2014cpp}, conceived to reduce memory consumption and computational time. The trees representing the individuals in the initial population and the functions used by geometric semantic operators are stored in memory, such that the subsequent individuals are composed of pointers to these structures. The semantics of the individuals is also stored in memory and used to compute the semantics of the offspring, reducing computational effort to calculate the fitness. 
However, the function represented by the individual is never truly built during the evolution. In order to obtain the symbolic function defined by the individual, we need to reconstruct it from the pointers and trees stored in memory, resulting in a function with size exponentially proportional to the number of generations.
Instead of using pointers and data structures, \citet{moraglio2014efficient} uses higher-order functions and memoization in his Python implementation, delegating the control to the compiler. The functions evolved are represented directly as compiled Python and still present exponential size when decompiled.

In the second group of works is the Subtree Semantic Geometric
Crossover (SSGX) operator, an approximation for the \textit{GSX}$_M$ that generates smaller individuals proposed by \citet{nguyen2016subtree}. Given two parent individuals, $p_1, p_2$, SSGX generates offspring by applying the \textit{GSX}$_M$ operator to the subtrees of $p_1$ and $p_2$ with semantics more similar to their respective parents, resulting in smaller functions. In addition, the method intercalates SSGX with the conventional subtree-swapping crossover \cite{koza1992genetic} during crossover operations, resulting in solutions around 29 orders of magnitude smaller than those generated by \textit{GSX}$_M$. However, although SSGX outperformed \textit{GSX}$_M$ in terms of test error in their experimental analysis, the configuration of the experiments can make the results inconclusive. This is because SSGX and \textit{GSX}$_M$ are tested using different mutation operators, which can be the responsible for the difference in the performance. In addition, SSGX is around three times more time consuming than \textit{GSX}$_M$. 

\citet{pawlak2017competent}, in turn, analyse a wide range of mutation and crossover operators under different metrics, including the ratio between the sizes of the offspring and their parents. The experimental analysis conducted with populations initialized using the ramped half-and-half method \cite{koza1992genetic} showed that \textit{GSM} generates offspring 5.16 times larger than their parents, on average, while the Tree Mutation (TM) \cite{koza1992genetic}, Competent Mutation (CM) \cite{pawlak2015thesis} and Semantically Driven Mutation (SDM) \cite{beadle2009semantically} result in offspring around 2.5 times larger than their parents. A similar experiment involving crossover operators showed that, on average, the Subtree-Swapping Crossover (SSX) \cite{koza1992genetic} and Semantically Driven Crossover (SDX) \cite{beadle2008semantically} generate offspring of the same size of their parents, and the Competent Crossover (CX) and \textit{GSX}$_E$ operators generate offspring 1.78 and 2.35 times larger than their parents, respectively. Notice, however, that TM and SSX are non-semantic operators, SDM and SDX are semantic but not geometric and CM and CX are approximately geometric semantic operators. 

The method proposed here does not fit in any of the approaches listed before.
It does not change the way \textit{GSM} or \textit{GSX} work and is able to generate exactly the same results that GSGP generates. 
It performs on-the-fly simplification of trees just after crossover and mutation operators are applied by taking advantage of the fact that the individuals are always linear combinations of trees. 
At the same time, its implementation does allow for efficient use of memory and computational time.

	\section{Methodology} \label{sec:methodology} 

\begin{figure}[t]
	\centering
	\includegraphics[width=\linewidth]{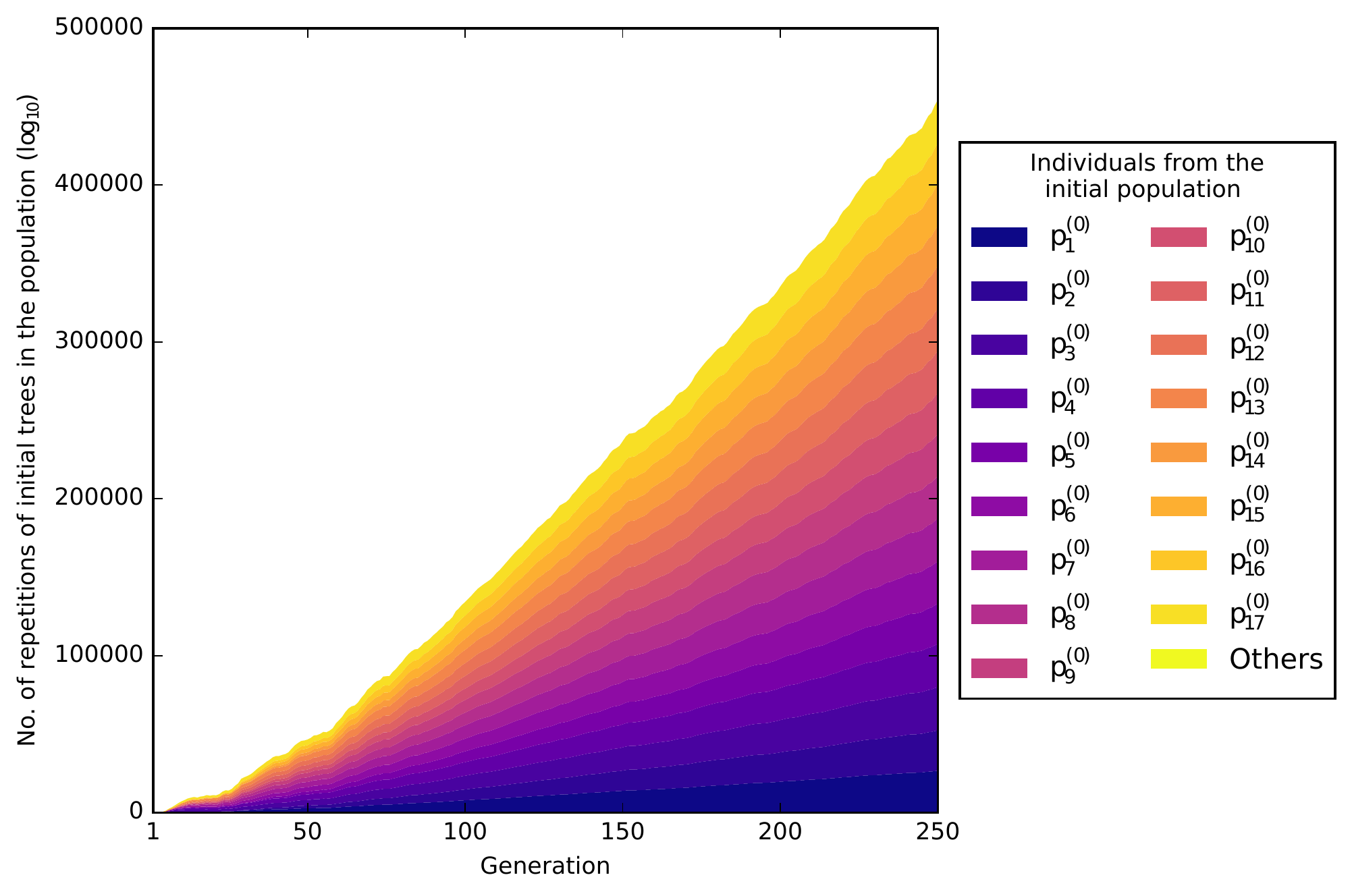}
	\caption{Frequency of appearance of initial population trees in the structure of individuals throughout generations.}
	\label{fig:frequency}
\end{figure}

By construction, GSGP operators keep the structure of the individuals being manipulated untouched, only adding other subtrees to them. Consequently, the structure of the individuals generated in the initial population is perpetuated through the whole evolution process, usually with multiple repetitions within the same individual. For instance, Figure \ref{fig:frequency} presents the frequency of the 1,000 individuals from the initial population composing the GSGP population throughout 250 generations.\footnote{This experiment was carried out with the same parameters adopted in Section \ref{sec:experiments} on the CCN dataset.} Notice that only 17 individuals generated in the initial population, represented as $p_1^{(0)}, p_2^{(0)}, \ldots, p_{17}^{(0)}$ in the legend, compose individuals in the later generations and all of them have a high frequency of appearance.

Given the large concentration of duplicated trees from the initial population within an individual, the size of the solutions evolved by GSGP can be drastically reduced by combining these repeated structures. 
The approach we propose, named Geometric Semantic Genetic Programming with Reduced trees (GSGP-Red), takes advantage of the repetition of functions to reduce the computational cost involved in the GSGP evolution. However, contrary to related methods from the literature, our approach combines repeated trees, effectively simplifying the individuals generated and reducing the size of the solutions dramatically.

\subsection{GSGP-Red}

GSGP-Red works by exploring the linear combinations of individuals performed by the geometric semantic mutation and crossover operators. Looking at Equation~\ref{eq:gsm}, observe that the \textit{GSM} operator generates a linear combination of the input parent and randomly generated trees. The \textit{GSX}$_E$ operator, in turn, generates a convex combination of the input parents. Thus, the evolution process performed by GSGP with these operators corresponds to recursively applying linear combinations to linear combinations of trees. GSGP-Red rewrites this recursion by expanding the terms and combining repeated structures. 

Let $P_0=\{p_1^{(0)}, p_2^{(0)}, \ldots, p_{|P_0|}^{(0)}\}$ be the initial GSGP population, $R=\{r_1, r_2, \ldots, r_{|R|}\}$ be the set of functions (trees) randomly generated by the \textit{GSM} operator during the evolutionary process. An individual $p_i$, from generation $g$, can be represented as a linear combination, defined by the dot product $C_i \cdot F_i$, where the first operand,
$C_i=[c_{i,1}, c_{i,2}, \ldots, c_{i,s_i}] \in \mathbb{R}^{s_i}$, is the set of coefficients multiplying the terms; the second operand, $F_i=[f_{i,1}, f_{i,2}, \ldots, f_{i,s_i}]^T \in \{P_0\cup R\}^{s_i}$, is the set of functions (trees) composing the individual, perpetuated from the initial population or generated by the \textit{GSM} operator; and $s_i$ is the number of distinct functions composing $p_i$. For $g=0$, the individual $p_i$ from the GSGP initial population can be described as in Equation~\ref{eq:GSGP3-Initial-Individual}. 
For $g > 0$, the individuals are recursively combined, as described in Equations~\ref{eq:GSGP3-GSM-expansion} and \ref{eq:GSGP3-GSX-expansion}.

\begin{equation}\label{eq:GSGP3-Initial-Individual}
	\begin{split}
	p_i^{(0)} & = 1 \times p_i^{(0)}  \\
              & = c_{i,1} \times f_{i,1} \\
              & = C_i \cdot F_i 
	\end{split}
\end{equation}

GSGP-Red applies two new steps to every new offspring generated by \textit{GSM} or \textit{GSX}$_E$, namely expansion and aggregation. During the expansion, GSGP-Red multiplies the coefficients from the geometric semantic operator by the randomly generated functions in \textit{GSM} offspring and adds all the terms, as presented in Equation \ref{eq:GSGP3-GSM-expansion}\footnote{For the sake of simplicity, we omit the input parameters of the functions.}. The same is done for the \textit{GSX}$_E$ offspring, but, in this case, the coefficients from the geometric semantic operator are multiplied by the coefficient vectors $C_i$ and $C_j$ from the parent individuals  $p_i$ and $p_j$, as presented in Equation \ref{eq:GSGP3-GSX-expansion}. At the end of the expansion stage, the offspring resulting from the \textit{GSM} and \textit{GSX}$_E$ operators---denoted as $p_{o_M}$ and $p_{o_X}$, respectively---consist of linear combinations, which can be rewritten as the dot products $C_{o_M}\cdot F_{o_M}$ and $C_{o_X}\cdot F_{o_X}$, respectively.

\begin{equation}\label{eq:GSGP3-GSM-expansion}
	\begin{split}
		\textit{GSM}(p_i, \delta) & = p_i + \delta\times(r_m-r_n)\\
									& = C_i \cdot F_i + \delta \times r_m  - \delta \times r_n\\
									& = c_{i,1}\times f_{i,1} + \ldots + c_{i,s_i}\times f_{i,s_i} \\
									&			\hspace{2cm}+\delta \times r_m - \delta \times r_n \\
									& = [c_{i,1}, \ldots, c_{i,s_i}, \delta, -\delta] \cdot [f_{i,1}, \ldots, f_{i,s_i}, r_m, r_n]^T \\
									& = C_{o_M}\cdot F_{o_M}
	\end{split}
\end{equation}

\begin{equation}\label{eq:GSGP3-GSX-expansion}
	\begin{split}
		\textit{GSX}_E(p_i, p_j)	& = k\times p_i + (1-k)\times p_j \\
										& = k\times (C_i \cdot F_i) + (1-k)\times (C_j \cdot F_j)\\
										& = k\times c_{i,1}\times f_{i,1} + \ldots + k\times c_{i,s_i}\times f_{i,s_i} \\
										& 			\hspace{1cm}+(1-k)\times c_{j,1}\times f_{j,1} + \ldots \\
										& 			\hspace{2cm}+ (1-k)\times c_{j,s_j}\times f_{j,s_j} \\
										& = [k\times c_{i,1}, \ldots, k\times c_{i,s_i}, \\
										&			\hspace{1cm} (1-k)\times c_{j,1},\ldots, (1-k)\times c_{j,s_j} ] \\
										& 			\hspace{1.5cm} \cdot [f_{i,1}, \ldots, f_{i,s_i}, f_{j,1}, \ldots, f_{j,s_j}]^T \\
										& = C_{o_X}\cdot F_{o_X}
	\end{split}
\end{equation}

During the aggregation stage, GSGP-Red combines functions appearing more than once in the list of functions from the resulting individual. Let $p_{new}=C_{new} \cdot F_{new}$ be an offspring resulting from the expansion step and $f_{rep}$ be a function appearing $l$ times in $F_{new}$, i.e.,  $f_{rep_1}, f_{rep_2}, \ldots, f_{rep_l}$, with the respective coefficients $c_{rep_1}, c_{rep_2}, \ldots, c_{rep_l}$ in $C_{new}$. The aggregation step keeps the first appearance of $f_{rep}$ ($f_{rep_1}$) in $F_{new}$ and its respective coefficient ($c_{rep_1}$) in $C_{new}$, removing all the other function occurrences and their respective coefficients---$ f_{rep_2}, \ldots, f_{rep_l}$ and $ c_{rep_2}, \ldots, c_{rep_l}$. 
For each coefficient removed, its value is added to the coefficient of the instance kept by the method, i.e., the value of $c_{rep_1}$ is updated to $\sum_{i=1}^{l} c_{rep_i}$, as they all come from a linear combination. The size of the individual---$s_{new}$---is also updated to reflect the removal of the repeated functions and their respective coefficients from the tree representation. 

Note that here we work with the \textit{GSX} defined w.r.t. the Euclidean distance.
The motivation for choosing \textit{GSX}$_E$ over \textit{GSX}$_M$ for GSGP-Red comes from the fact that \textit{GSX}$_M$ multiplies the parents by a randomly generated function---contrary to the linear combination performed by \textit{GSX}$_E$---which would imply in additional complexity in time and space to store and manipulate a function instead of a constant. Notice that the usage of the \textit{GSX}$_E$ over the \textit{GSX}$_M$ or vice versa is an open discussion in the literature, with some works defending the usage of \textit{GSX}$_M$, given empirical analysis \cite{nguyen2016subtree}, and others defending the usage of \textit{GSX}$_E$, given its progression properties \cite{pawlak2015progress}.

Next, we illustrate the expansion and aggregation operators. Consider a initial population with individuals $P=\{p_1^{(0)}, p_2^{(0)}, p_3^{(0)} \}$, where

\begin{align}
	p_1^{(0)} & = x_1/x_2 & = \ & 1\times (x_1/x_2) & = \ & c_{1,1}\times f_{1,1}~,\\
    p_2^{(0)} & = x_2+0.4 & = \ & 1\times (x_2+0.4) & = \ & c_{2,1}\times f_{2,1}~,\\ 
    p_3^{(0)} & = x_1-0.6 & = \ & 1\times (x_1-0.6) & = \ & c_{3,1}\times f_{3,1}~.
\end{align}

\noindent
Crossing over $p_2^{(0)}$ with $p_3^{(0)}$ (Equation~\ref{eq:example-g1-p1}) and mutating $p_1^{(0)}$ (Equation~\ref{eq:example-g1-p2}) and $p_3^{(0)}$ (Equation~\ref{eq:example-g1-p3}), with a mutation step arbitrarily set to 0.1, results in three new individuals, $p_1^{(1)}, p_2^{(1)}$ and $p_3^{(1)}$, respectively, composing the population of the next generation.

 \begin{align}
     \begin{split}
     \label{eq:example-g1-p1}
     p_1^{(1)} & = 0.3\times p_2^{(0)}+(1-0.3)\times p_3^{(0)}\\
     & = 0.3\cdot [1\times (x_2+0.4)]+(1-0.3)\times[1\times (x_1-0.6)] \\
                & = [0.3\times (x_2+0.4)] + [0.7\times (x_1-0.6)]\\
                & = c_{1,1}\times f_{1,1} + c_{1,2}\times f_{1,2}
     \end{split} \\
     \begin{split}
     \label{eq:example-g1-p2}
     p_2^{(1)} & = p_1^{(0)} + 0.1 \times [(x_1)-(2\times x_2)]\\
      & = (x_1/x_2) + 0.1 \times [(x_1)-(2\times x_2)]\\     
                & = [1\times (x_1/x_2)] + [0.1\times (x_1)] + [-0.1\times(2\times x_2)]\\
                & = c_{2,1}\times f_{2,1}+c_{2,2}\times f_{2,2}+c_{2,3}\times f_{2,3}
     \end{split} \\
     \begin{split}
     \label{eq:example-g1-p3}
       p_3^{(1)} & =  p_3^{(0)} + 0.1 \times [(x_1-0.6)- (x_1\times x_2)]\\
                & =  x_1-0.6 + 0.1 \times [(x_1-0.6)- (x_1\times x_2)]\\
                & = [1 \times \mathbf{(x_1-0.6)}] + [0.1 \times \mathbf{(x_1-0.6)}] \\
                & \hspace{2cm} + [-0.1 \times (x_1\times x_2)]\\
                & = [(1+0.1) \times \mathbf{(x_1-0.6)}] + [-0.1 \times (x_1\times x_2)]\\
                & = c_{3,1}\times f_{3,1}+c_{3,2}\times f_{3,2}
     \end{split}
  \end{align}

In this example, the random constant used by crossover in Eq. \ref{eq:example-g1-p1} is equal to $0.3$ and the two functions randomly generated by the mutation operator are $x_1$ and $2\times x_2$ in Eq. \ref{eq:example-g1-p2} and $x_1-0.6$ and $x_1\times x_2$ in Eq. \ref{eq:example-g1-p3}. Notice that one of the random functions generated in Eq. \ref{eq:example-g1-p3} is equal to the function represented by the parent individual---both presented in bold. These functions are then combined in a single function and the coefficients are summed up, generating a smaller individual. 

\subsection{Implementation Details}

The implementation employed in the experimental analysis of Section \ref{sec:experiments} iterates through the functions composing the new individual, performing the expansion and aggregation steps sequentially. In addition, it keeps a hash table for each individual to store the trees and their respective coefficients uniquely---indexed by the symbolic expression represented by the tree---speeding up the process of aggregating the functions composing an individual. This is the only form of representation that is necessary for each individual, with no direct pointers to its parents, who are implicitly stored amongst the expanded and then aggregated set of functions and their respective coefficients, unlike previous implementations \cite{vanneschi2013new,castelli2014cpp}. The code is available for download\footnote{\url{https://github.com/laic-ufmg/GSGP-Red}} or can also be directly executed from the Lemonade\footnote{\url{https://demo.ctweb.inweb.org.br}} data analysis platform.

\begin{figure}[t]
	\centering
	\includegraphics[scale=0.4235]{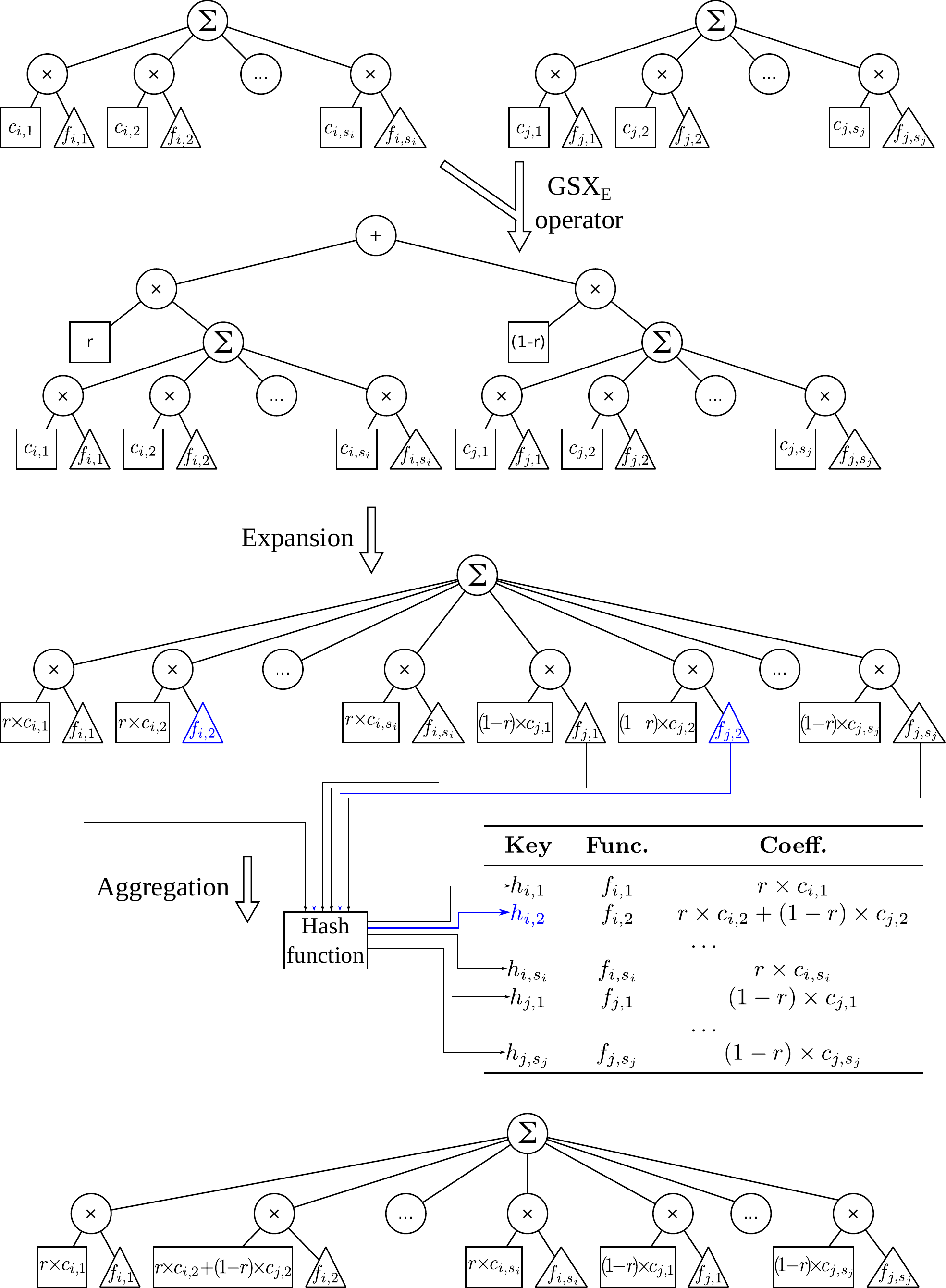}
	\caption{Expansion and aggregation after applying the \textit{GSX}$_E$ operator.}
	\label{fig:gsgp3-xover}
\end{figure}

\begin{figure}[t]
	\centering
	\includegraphics[scale=0.4235]{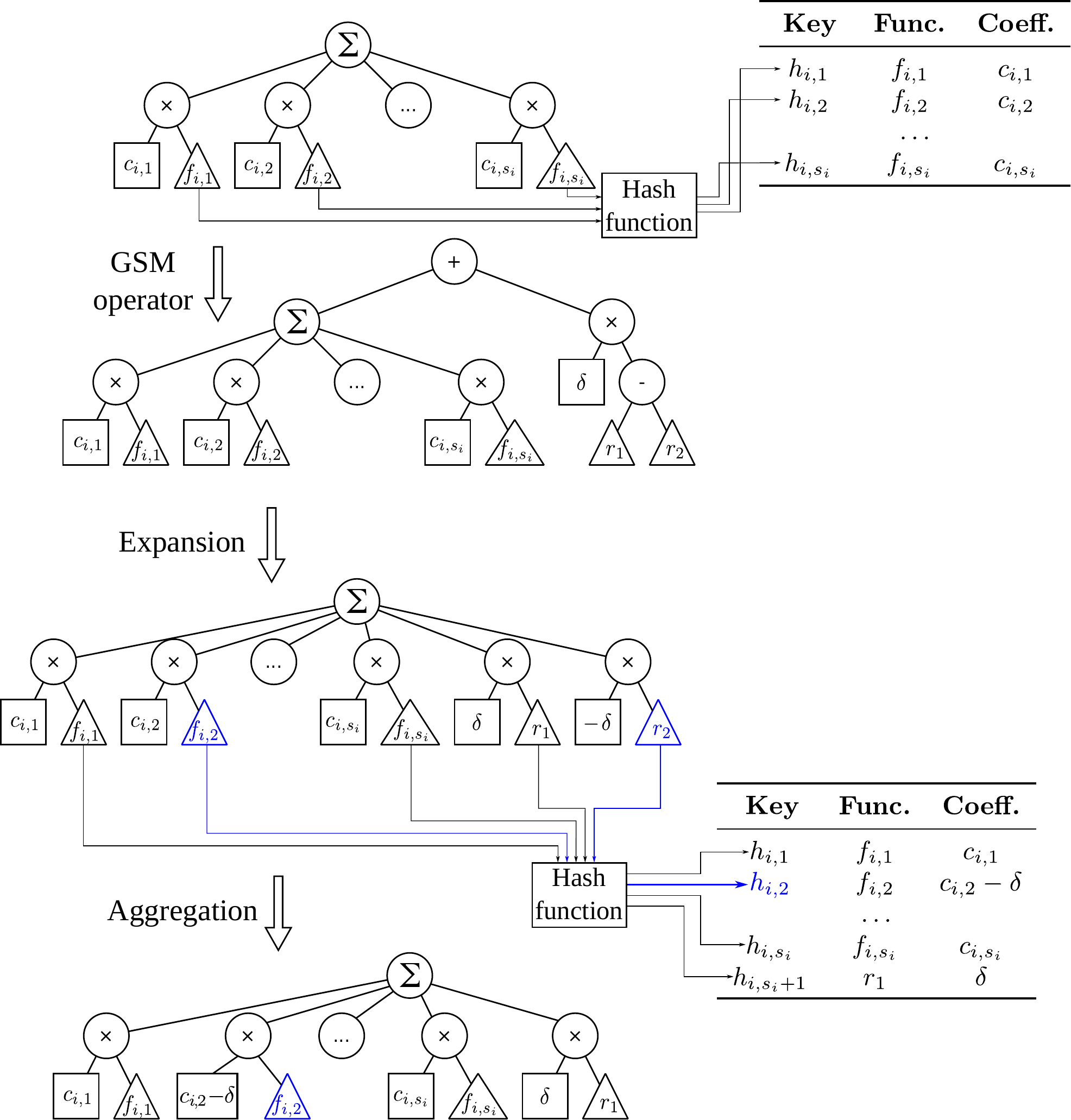}
	\caption{Expansion and aggregation after applying the \textit{GSM} operator.}
	\label{fig:gsgp3-mutation}
\end{figure}

Figures \ref{fig:gsgp3-xover} and \ref{fig:gsgp3-mutation} depict the functioning of GSGP-Red with the hash tables used in our implementation. The hash function maps the functions composing the individual to a hash key, used to index the table. Figure~\ref{fig:gsgp3-xover} presents a \textit{GSX}$_E$ offspring transformed by GSGP-Red---the hash tables of the parents are omitted in the figure. The expansion step results in two repeated trees---represented by $f_{i,2}$ and $f_{j,2}$, in blue in the figure---which are combined in the aggregation step---notice the sum of the coefficients $r\times c_{i,2}$ and $(1-r)\times c_{j,2}$ and the hash function resulting in only one blue arrow.

Fig. \ref{fig:gsgp3-mutation}, on the other hand, presents the GSGP-Red procedure applied to a \textit{GSM} offspring. In our example, the mutation operator generates a tree---$r_2$---equivalent to a function composing the parent individual---$f_{i,2}$---resulting in the same hash index---presented in blue. During the aggregation, the coefficients of these functions are combined---$c_{i,2}+(-\delta)$---when the hash table is updated. Notice that, although infrequent, it is possible for the \textit{GSM} to generate a tree already generated somewhere else during the evolution, given the finite number of possible combinations of functions and terminals.

	\section{Experimental Analysis}\label{sec:experiments}

In this section we present an experimental analysis of the performance of GSGP-Red when applied to 12 real-world datasets. 
The datasets are described in Table \ref{table:datasets}, which shows their number of attributes (\# of attrs) and number of instances (\# of instances). 
GSGP-Red results are compared to the results obtained by GSGP and the canonical GP \cite{banzhaf1998genetic} in terms of training and test Root Mean Square Error (RMSE), size of the functions---given by the number of tree nodes---and computational time expended by the methods. 

Given the non-deterministic nature of the methods, each experiment was repeated 30 times---6 times for each fold in a 5-fold cross-validation procedure. In order to validate the results, we performed Wilcoxon signed-rank tests \cite{demvsar2006statistical,sheskin2003handbook} with a confidence level of 95\%, under the null hypothesis that GSGP-Red performance---in terms of test RMSE, solution size and computational time---is equal to the performance of the other methods for each dataset. 

\begin{table}[t]
	\small
	\rowcolors{2}{gray!25}{white}
	\caption{Datasets used in the experiments.}
	\begin{center}
		\begin{tabular}{llrrr}
			\toprule
			\textbf{Abbr.} & \textbf{Dataset} & \textbf{\# of attrs} & \textbf{\# of instances} & \textbf{Source}\\
			\cmidrule(r{0em}){1-5}
			air & Airfoil & 6 & 1503 & \cite{albinati2015effect,lichman2015}\\
			ccn & CCN & 123 & 1994 & \cite{lichman2015}\\
			ccun & CCUN & 125 & 1994 & \cite{lichman2015}\\
			con & Concrete  & 9 & 1030 & \cite{albinati2015effect,lichman2015}\\
			eneC & Energy Cooling & 9 & 768 & \cite{albinati2015effect,lichman2015}\\
			eneH & Energy Heating & 9 & 768 & \cite{albinati2015effect,lichman2015}\\
			par & Parkinsons & 19 & 5875 & \cite{lichman2015}\\      
			ppb & PPB & 627 & 131 & \cite{albinati2015effect}\\
			tow & Tower & 26 & 4999 & \cite{albinati2015effect}\\
			wineR & Wine Red & 12 & 1599 & \cite{albinati2015effect,lichman2015}\\
			wineW & Wine White & 12 & 4898 & \cite{albinati2015effect,lichman2015}\\
			yac & Yacht & 7 & 308 & \cite{albinati2015effect,lichman2015}\\
			\midrule[\heavyrulewidth] 
		\end{tabular}
	\end{center}

	\label{table:datasets}
\end{table}

\subsection{GP and GSGP settings}

GP and GSGP were run with a population of 1,000 individuals evolved for 250 generations with tournament selection of size 7 and 10, respectively. 
The terminal set comprised the variables of the problem and constant values uniformly picked from $[-1,1]$, described by \citet{koza1992genetic} as \textit{ephemeral random constants} (ERC). 
The function set included three binary arithmetic operators ($+, -, \times$) and the analytic quotient (AQ) \cite{ni2013use}, which has the general properties of division but without discontinuity, given by:

\begin{equation}
\label{eq:aq}
\textit{AQ}(a,b)=\frac{a}{\sqrt{1+b^2}}
\end{equation}

The GP method employed the canonical crossover and mutation operators \cite{koza1992genetic} with probabilities $0.9$ and $0.1$, respectively. GSGP employed \textit{GSX}$_E$ and \textit{GSM} operators, both with probability $0.5$, as presented in \cite{castelli2014cpp}, but without the logistic function to bound the outputs of the randomly generated functions. The grow method \cite{koza1992genetic} was adopted to generate the random functions within the geometric semantic crossover and mutation operators, and the ramped half-and-half method \cite{koza1992genetic} to generate the initial population, both with a maximum individual depth equal to 6. Following the work from \citet{albinati2014effect}, the mutation step adopted by the geometric semantic mutation operator was defined as 10\% of the standard deviation of the target output vector given by the training data.
Both methods used the RMSE calculated over the obtained and expected output values for the training set.
The same parameters adopted for GSGP were used in GSGP-Red experiments.

\subsection{Experimental Analysis}

\begin{table}[t]
    \centering
    \caption{Median RMSE of the best individual for training and test sets for GP and GSGP/GSGP-Red. The symbol \up (\down) indicates that the performance of GSGP-Red was better (worse) than the performance of GP.}
    \begin{tabular}{cccc>{\hspace*{-2.7mm}}c} 
        \toprule
        \textbf{Dataset} & \textbf{RMSE} & \textbf{GSGP/GSGP-Red} & \multicolumn{2}{c}{\textbf{GP}} \\
        \midrule
        \multirow{2}{*}{air} & \multicolumn{1}{c}{Training} & \multicolumn{1}{c}{11.783} & \multicolumn{1}{c}{17.353} & \\
        & \multicolumn{1}{c}{\cellcolor{gray!25}Test} & \multicolumn{1}{c}{\cellcolor{gray!25}11.280} & \multicolumn{1}{c}{\cellcolor{gray!25}17.612} & {\cellcolor{gray!25}\up} \\
        
        
        \cellcolor{gray!25} & \multicolumn{1}{c}{Training} & \multicolumn{1}{c}{0.128} & \multicolumn{1}{c}{0.145} & \\
        \multirow{-2}{*}{\cellcolor{gray!25}ccn} & \multicolumn{1}{c}{\cellcolor{gray!25}Test} & \multicolumn{1}{c}{\cellcolor{gray!25}0.138} & \multicolumn{1}{c}{\cellcolor{gray!25}0.149} & {\cellcolor{gray!25}\up} \\
        
        
        \multirow{2}{*}{ccun} & \multicolumn{1}{c}{Training} & \multicolumn{1}{c}{377.616} & \multicolumn{1}{c}{386.203} & \\
        & \multicolumn{1}{c}{\cellcolor{gray!25}Test} & \multicolumn{1}{c}{\cellcolor{gray!25}405.463} & \multicolumn{1}{c}{\cellcolor{gray!25}396.308} & {\cellcolor{gray!25}\eq} \\
        
        
        \cellcolor{gray!25} & \multicolumn{1}{c}{Training} & \multicolumn{1}{c}{8.510} & \multicolumn{1}{c}{9.352} & \\
        \multirow{-2}{*}{\cellcolor{gray!25}con} & \multicolumn{1}{c}{\cellcolor{gray!25}Test} & \multicolumn{1}{c}{\cellcolor{gray!25}8.886} & \multicolumn{1}{c}{\cellcolor{gray!25}9.750} & {\cellcolor{gray!25}\up} \\
        
        
        \multirow{2}{*}{eneC} & \multicolumn{1}{c}{Training} & \multicolumn{1}{c}{3.114} & \multicolumn{1}{c}{3.364} & \\
        & \multicolumn{1}{c}{\cellcolor{gray!25}Test} & \multicolumn{1}{c}{\cellcolor{gray!25}3.129} & \multicolumn{1}{c}{\cellcolor{gray!25}3.455} & {\cellcolor{gray!25}\up} \\
        
        
        \cellcolor{gray!25} & \multicolumn{1}{c}{Training} & \multicolumn{1}{c}{2.677} & \multicolumn{1}{c}{2.973} & \\
        \multirow{-2}{*}{\cellcolor{gray!25}eneH} & \multicolumn{1}{c}{\cellcolor{gray!25}Test} & \multicolumn{1}{c}{\cellcolor{gray!25}2.739} & \multicolumn{1}{c}{\cellcolor{gray!25}3.101} & {\cellcolor{gray!25}\up} \\
        
        
        \multirow{2}{*}{par} & \multicolumn{1}{c}{Training} & \multicolumn{1}{c}{9.812} & \multicolumn{1}{c}{9.955} & \\
        & \multicolumn{1}{c}{\cellcolor{gray!25}Test} & \multicolumn{1}{c}{\cellcolor{gray!25}9.868} & \multicolumn{1}{c}{\cellcolor{gray!25}9.995} & {\cellcolor{gray!25}\up} \\
        
        
        \cellcolor{gray!25} & \multicolumn{1}{c}{Training} & \multicolumn{1}{c}{14.870} & \multicolumn{1}{c}{27.644} & \\
        \multirow{-2}{*}{\cellcolor{gray!25}ppb} & \multicolumn{1}{c}{\cellcolor{gray!25}Test} & \multicolumn{1}{c}{\cellcolor{gray!25}29.647} & \multicolumn{1}{c}{\cellcolor{gray!25}28.542} & {\cellcolor{gray!25}\eq} \\
        
        
        \multirow{2}{*}{tow} & \multicolumn{1}{c}{Training} & \multicolumn{1}{c}{46.634} & \multicolumn{1}{c}{50.050} & \\
        & \multicolumn{1}{c}{\cellcolor{gray!25}Test} & \multicolumn{1}{c}{\cellcolor{gray!25}46.533} & \multicolumn{1}{c}{\cellcolor{gray!25}50.155} & {\cellcolor{gray!25}\up} \\
        
        
        \cellcolor{gray!25} & \multicolumn{1}{c}{Training} & \multicolumn{1}{c}{0.632} & \multicolumn{1}{c}{0.657} & \\
        \multirow{-2}{*}{\cellcolor{gray!25}wineR} & \multicolumn{1}{c}{\cellcolor{gray!25}Test} & \multicolumn{1}{c}{\cellcolor{gray!25}0.636} & \multicolumn{1}{c}{\cellcolor{gray!25}0.652} & {\cellcolor{gray!25}\up} \\
        
        
        \multirow{2}{*}{wineW} & \multicolumn{1}{c}{Training} & \multicolumn{1}{c}{0.729} & \multicolumn{1}{c}{0.756} & \\
        & \multicolumn{1}{c}{\cellcolor{gray!25}Test} & \multicolumn{1}{c}{\cellcolor{gray!25}0.735} & \multicolumn{1}{c}{\cellcolor{gray!25}0.766} & {\cellcolor{gray!25}\up} \\
        
        
        \cellcolor{gray!25} & \multicolumn{1}{c}{Training} & \multicolumn{1}{c}{6.529} & \multicolumn{1}{c}{3.413} & \\
        \multirow{-2}{*}{\cellcolor{gray!25}yac} & \multicolumn{1}{c}{\cellcolor{gray!25}Test} & \multicolumn{1}{c}{\cellcolor{gray!25}6.437} & \multicolumn{1}{c}{\cellcolor{gray!25}3.541} & {\cellcolor{gray!25}\down} \\
        
        
        \midrule[\heavyrulewidth] 
    \end{tabular}

    \label{table:fitness}
\end{table}

Table \ref{table:fitness} presents the RMSE obtained by each method on the 12 datasets. The symbol \eq ~indicates the null hypothesis (GSGP-Red performance is equal to the performance of other methods) was not discarded and the symbol \up (\down) indicates that the performance of GSGP-Red was better (worse) than the performance of the GP.
Recall that the results of GSGP/GSGP-Red are the same (see Section \ref{sec:methodology} for details), as the proposed method generates a solution equivalent to the one produced by GSGP.
According to the outcomes of the statistical tests regarding the test RMSE, GP is better than GSGP in the Yacht dataset, and the results have no statistical difference for datasets CCUN and PPB. 
In the other 9 cases, GSGP/GSGP-Red is better than GP, which motivates the construction of methods such as GSGP-Red.

The previous results confirm that the solutions generated by GSGP-Red are equivalent to those generated by GSGP and, in most cases, better than the solutions produced by the canonical GP.
However, the results that show the main contribution of the proposed method are listed in Table \ref{table:size}, where we show the median number of nodes in the best individuals of GSGP-Red, GSGP and GP. Again,
the symbol \eq ~indicates the null hypothesis (GSGP-Red size is equal to the size of other methods) was not discarded and the symbol \up (\down) indicates that the performance of GSGP-Red was better (worse) than the performance of the method indicated by the column (GSGP or GP).

Note that GSGP-Red individuals are always much smaller than those generated by GSGP.
By calculating the reduction in size when comparing GSGP to GSGP-Red, solutions from the latter are, on average, 58 orders of magnitude smaller. The maximum reduction in size occurred in CCUN (64 order of magnitude) and the minimum reduction occurred in the Parkinsons dataset (45 orders of magnitude). 
It is important to point out that the function sizes are still substantially bigger than the ones generated by GP, but without forgetting that the RMSE results for GSGP are still, in general, superior. As discussed later, we believe that the functions generated by GSGP-Red can be further reduced using other techniques, such as algebraic simplification.

\begin{table}[t]
	\centering
	\rowcolors{2}{gray!25}{white}
	\caption{Median size of the best individual (in number of nodes) for GSGP-Red, GSGP and GP. The symbol \up (\down) indicates that the performance of GSGP-Red was better (worse) than the performance of the method indicated in the column.}
	\begin{tabular}{lcc>{\hspace*{-3mm}}cc>{\hspace*{-3mm}}c}
		\toprule
		\textbf{Dataset} & \textbf{GSGP-Red} & \multicolumn{2}{c}{\textbf{GSGP}} & \multicolumn{2}{c}{\textbf{GP}} \\
		\midrule
		{air} & 33,353 & 1.72e+50 & ~\up & 86 & ~\down \\
		\addlinespace[0.1cm]
		{ccn} & 22,819 & 7.25e+58 & ~\up & 40 & ~\down \\
		\addlinespace[0.1cm]
		{ccun} & 3,373 & 2.33e+67 & ~\up & 45 & ~\down \\
		\addlinespace[0.1cm]
		{con} & 6,007 & 1.23e+65 & ~\up & 43 & ~\down \\
		\addlinespace[0.1cm]
		{eneC} & 6,584 & 1.08e+65 & ~\up & 64 & ~\down \\
		\addlinespace[0.1cm]
		{eneH} & 6,881 & 1.19e+65 & ~\up & 67 & ~\down \\
		\addlinespace[0.1cm]
		{par} & 34,386 & 1.88e+49 & ~\up & 81 & ~\down \\
		\addlinespace[0.1cm]
		{ppb} & 12,185 & 2.29e+64 & ~\up & 61 & ~\down \\
		\addlinespace[0.1cm]
		{tow} & 4,843 & 7.34e+66 & ~\up & 49 & ~\down \\
		\addlinespace[0.1cm]
		{wineR} & 9,220 & 3.53e+64 & ~\up & 49 & ~\down \\
		\addlinespace[0.1cm]
		{wineW} & 9,983 & 2.20e+64 & ~\up & 45 & ~\down \\
		\addlinespace[0.1cm]
		{yac} & 13,706 & 1.23e+61 & ~\up & 62 & ~\down \\
		\midrule[\heavyrulewidth] 
	\end{tabular}

	\label{table:size}
\end{table}

\subsection{Run-time Analysis}

In order to analyse to what extent the application of the expansion and aggregation processes increase GSGP computational cost, we compare the median time spent by GSGP-Red and the canonical versions of GP and GSGP to generate regression models for our testbed, including both training and test stages. The results of this analysis, shown in Table \ref{table:time}, indicate that the running times for GSGP-Red are, in general, higher than those presented by GSGP, but there are exceptions. For some datasets, GSGP-Red was not only faster than GP but also faster than GSGP itself. This can be explained by the fact that, when running GSGP-Red, we do not need to calculate the test fitness for every created individual. This was mandatory in the GSGP implementation proposed by Vanneschi \cite{vanneschi2013new}, for example, as the best individual could not be easily reconstructed at the end of the evolution process, and values of RMSE for training and test were computed during evolution. This is not the case for GSGP-Red, which can easily store the solution generated for later use in new data. Hence, GSGP-Red only evaluates the test fitness of the best overall individual. For some datasets, removing these operations make the total runtime decrease considerably, making GSGP-Red adoption even more appealing.

GSGP-Red was faster than GSGP in three datasets: CCUN, Tower and Wine White, being statistically worse in all the remaining. However, the fact that GSGP-Red takes longer to run does not indicate a lack of efficiency or some fundamental problem compromising its application since, in absolute terms, the difference between the two methods is still small. On the other hand, when compared to GP, GSGP-Red is faster in 10 out of 12 datasets, with execution times, on average, 35\% faster. In conclusion, GSGP-Red is overall slower than GSGP but is still efficient and yet better than GP in terms of RMSE and execution time.

\begin{table}[t]
	\centering
	\caption{Median of the execution time (in seconds) of GSGP-Red, GSGP and GP. The symbol \up (\down) indicates that the performance of GSGP-Red was better (worse) than the performance of the method indicated in the column.}
	\rowcolors{2}{gray!25}{white}
	\begin{tabular}{lcc>{\hspace*{-3mm}}cc>{\hspace*{-3mm}}c} 
		\toprule
		\textbf{Dataset} & \textbf{GSGP-Red} & \multicolumn{2}{c}{\textbf{GSGP}} & \multicolumn{2}{c}{\textbf{GP}} \\
		\midrule
		{air} & 113.28 & 48.32 & \down & 202.13 & \up \\
		\addlinespace[0.1cm]
		{ccn} & 96.10 & 63.35 & \down & 154.75 & \up \\
		\addlinespace[0.1cm]
		{ccun} & 53.15 & 63.62 & \up & 148.98 & \up \\
		\addlinespace[0.1cm]
		{con} & 39.19 & 35.17 & \down & 77.85 & \up \\
		\addlinespace[0.1cm]
		{eneC} & 36.24 & 28.99 & \down & 78.65 & \up \\
		\addlinespace[0.1cm]
		{eneH} & 37.85 & 27.92 & \down & 86.57 & \up \\
		\addlinespace[0.1cm]
		{par} & 191.93 & 165.36 & \down & 767.46 & \up \\
		\addlinespace[0.1cm]
		{ppb} & 37.58 & 10.74 & \down & 15.57 & \down \\
		\addlinespace[0.1cm]
		{tow} & 110.60 & 136.90 & \up & 486.54 & \up \\
		\addlinespace[0.1cm]
		{wineR} & 55.03 & 48.32 & \down & 128.04 & \up \\
		\addlinespace[0.1cm]
		{wineW} & 127.24 & 140.66 & \up & 378.36 & \up \\
		\addlinespace[0.1cm]
		{yac} & 39.19 & 15.03 & \down & 32.35 & \down \\
		\midrule[\heavyrulewidth] 
	\end{tabular}
	\label{table:time}
\end{table}

	\section{Conclusions and Future Work}

This paper presented Geometric Semantic Genetic Programming with Reduced trees (GSGP-Red), a new method that solves the exponential growth of GSGP solutions with the number of generations for symbolic regression problems. The method expands the functions representing the individuals into linear combinations, and then aggregates the repeated structures. This process results in functions many times smaller than those generated by GSGP.

An experimental analysis was performed in a testbed composed of 12 real-world datasets in order to compare GSGP-Red with its predecessor and with GP. Results showed that the new method is capable of generating solutions equivalent to those generated by GSGP in terms of error, but 58 orders of magnitude smaller, on average., in terms of size (number of tree nodes). In addition, an analysis of the execution time revealed that GSGP-red, although slower than GSGP on average, can also perform the search faster than GSGP, depending on the dataset.

Potential future developments include simplifying the solutions using computer algebra systems \cite{moraglio2012geometric} and integrating approximated geometric semantic operators---e.g., the competent mutation and crossover operators from \citet{pawlak2015thesis}---to GSGP-Red, in order to reduce even further the size of the solutions generated.

Compiling all these ideas into a single framework seems a promising direction to make the readability and degree of understanding of the GSGP solutions closer to those of models generated by GP.

	\begin{acks}
		
		This work was partially supported by the following Brazilian Research Support Agencies: CNPq, FAPEMIG, CAPES. The authors' work has also been partially funded by the EUBra-BIGSEA project by the European Commission under the Cooperation Programme (MCTI/RNP 3rd Coordinated Call), Horizon 2020 grant agreement 690116. Finally, we would like to thank Gabriel Coutinho for his valuable insights on the mathematical aspects of the problem in question.
        
	\end{acks}
	
	\bibliographystyle{ACM-Reference-Format}
	\bibliography{references}
	
\end{document}